\documentclass[11pt,a4paper]{article}
\usepackage[hyperref]{deepcuts}
\usepackage{times}
\usepackage{latexsym}
\usepackage{fontawesome}
\usepackage{mathtools}
\usepackage{fancyhdr}
\usepackage{subfig}

\usepackage{xurl}

\title{DeepCuts: Single-Shot Interpretability based Pruning for BERT }

\author{Jasdeep Singh Grover* \\
  \\\And
  Bhavesh Gawri* \\
  {\tt \{jasgrover,rmanku,bgawri\}@cs.stonybrook.edu} \\\And
  Ruskin Raj Manku* \\
  }

\begin{document}

\maketitle
\footnotetext[1]{* indicates equal contributions. \\Code available at: \url{https://github.com/RuskinManku/DeepCuts}}%

\begin{abstract}

As language models have grown in parameters and layers, it has become much harder to train and infer with them on single GPUs. This is severely restricting the availability of large language models such as GPT-3, BERT-Large, and many others. A common technique to solve this problem is pruning the network architecture by removing transformer heads, fully-connected weights, and other modules. The main challenge is to discern the important parameters from the less important ones. Our goal is to find strong metrics for identifying such parameters. We thus propose two strategies: Cam-Cut based on the GradCAM interpretations, and Smooth-Cut based on the SmoothGrad, for calculating the importance scores. Through this work, we show that our scoring functions are able to assign more relevant task-based scores to the network parameters, and thus both our pruning approaches significantly outperform the standard weight and gradient-based strategies, especially at higher compression ratios in BERT-based models. We also analyze our pruning masks and find them to be significantly different from the ones obtained using standard metrics.

\end{abstract}
% \pagestyle{fancy}
% \fancyfoot[L]{Testing}

\section{Introduction}

\begin{figure}
    \centering
    \includegraphics[width=0.5\textwidth]{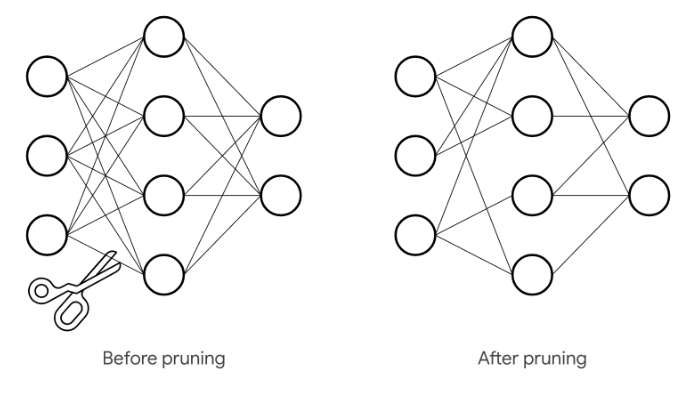}
    \caption{A schematic representation of pruning \cite{tfprune}.}
    \label{fig:pruning}
\end{figure}

It has been extremely challenging to train or even at times fine-tune large language models like BERT-Large, RoBERTa-Large, BART-large, and many others on single GPU devices. Simultaneously, many applications require model deployment on less powerful devices like mobile devices and in federated learning over multiple edge devices which cannot be supported for very large language models. Thus reducing the model size and retaining as much performance as possible is essential for these applications and less resourceful users. Hence, our objective is to identify strategies and scoring functions for discerning essential weights from the less important ones with special consideration to the dataset and task in hand. This is similar to the schematic shown in \ref{fig:pruning}. We thus perform single-shot pruning to compare the performance of our metrics with standard weight and gradient-based methods. We believe interpretability techniques developed for Explainable ML like GradCAM \cite{selvaraju2017grad} and SmoothGrad \cite{smilkov2017smoothgrad} can provide significant bases for weight identification and thus can be repurposed for pruning the weights of large language models. Recent works have leveraged the Lottery Ticket Hypothesis (LTH) \cite{frankle2018lottery,prasanna2020bert} which states that for every large model, we can identify a smaller sub-model that when trained from the same initialization will perform similarly or even at times better than the original model. But it is very difficult to efficiently identify these sub-models especially when we want them to be much smaller than the original model. 

% Paragraph 2: What are the broad approaches that are used to address this problem? Please cite papers for each of these approaches.\\

\subsection{Related Works}

The pruning problem is mainly tackled using two broad categories of approaches \cite{blalock2020state}. These include unstructured pruning \cite{han2015deep, gale2019state}, and structure-based pruning \cite{voita2019analyzing,zhang2021know,perez2022topological}. Unstructured pruning involves scoring individual prunable structures for removing individual parameters from modules whereas structured pruning includes removing complete modules from the model like complete attention heads and MLP layers from transformer architectures. Other pruning methods use probability-guided pruning where probabilities are assigned to parameters based on the given scoring function and the pruning is done by sampling from this distribution \cite{wang2017structured}. There are also mask learning approaches that directly learn pruning masks during the training methods and use them to prune the weights\cite{louizos2017learning}. Many methods in tabular machine learning also have conditions for independence of feature interactions, which also leads to smaller models used especially in financial applications like GAMI-Net \cite{yang2021gami}. Many works \cite{han2015deep,frankle2018lottery,sanh2020movement,prasanna2020bert} follow an iterative pruning and fine-tuning strategy where a certain percentage of the model is pruned in every cycle followed by fine-tuning. Other works explore single-shot pruning, i.e. pruning and training the network only once \cite{lee2018snip}. 

Many of the existing works use magnitude-based approaches that use absolute network weights to directly assign scores to the parameters, i.e. the higher the absolute value of the parameter, the higher its importance, and the less likely it is to be pruned \cite{frankle2018lottery, prasanna2020bert}. The main issue with this approach is that if many low-value parameters belong to any of the earlier layers in the network, they would be pruned and most of the information flowing through the later layers would be cut off. This issue is resolved in the existing literature \cite{lee2020layer} by applying a cap on the max number of parameters that can be pruned in each layer. These works focus on the magnitude of the parameters which by themselves do not carry all the task-relevant information in hand and thus do not perform well. Many recent works \cite{sanh2020movement, zhang2021know, kurtic2022optimal} explore the combinations of gradients and parameter values. These works explore different variants of the gradients in the scoring functions for instance some use the absolute value of gradients \cite{zhang2021know} while others consider the directions of gradients as well \cite{sanh2020movement}. Some works \cite{yu2022hessian, kurtic2022optimal} also explore higher-order gradients such as Hessian but their calculations could become compute and memory intensive.

Machine Learning Interpretability methods like GradCAM\cite{selvaraju2017grad}, Integrated Gradients \cite{sundararajan2017axiomatic}, and Smooth Grad \cite{smilkov2017smoothgrad} have been used to understand models in the recent past. They obtain saliency maps for computer vision and language applications \cite{dai2021knowledge} using methods like LIME \cite{ribeiro2016should}, GradCAM, SmoothGrad, and others which have been used to attribute significance to a given pixel or token for a given output. We believe that we can leverage these techniques to do the same for the parameters of the architecture that can serve as a score for pruning. Pruning and interpretability face similar challenges. In interpretability, noisy gradients have always been a challenge, and the same is also observed in pruning. Similarly, excluding the magnitude of input hinders interpretability and the same is true for parameter magnitude in pruning. However, these challenges have been addressed through SmoothGrad, and Inverted Representations \cite{mahendran2015understanding} in interpretability. Simultaneously, using a combination of gradients and input image as a product has been a very common approach in the interpretability literature to identify important regions in input that contribute the most to generating the given output. The same practice is being used in pruning using the product of weight magnitude and gradient as a scoring function. Thus, we believe ideas like SmoothGrad and GradCAM can provide a very intuitive and effective foundation for scoring functions for pruning.

% \begin{figure}[]
% 	\centering
% 	\begin{subfigure}[]{\includegraphics[width=0.30\textwidth]{./images/dog.png}}\end{subfigure}
% 	% \begin{subfigure}[]{\includegraphics[width=0.30\textwidth]{./images/dog.png}}\end{subfigure}
% 	\caption{\small(a) fjsd}
% \end{figure}

%Paragraph 5: Briefly mention how you tested these ideas. Mention the system that you are starting as your baseline and then say what you are doing to this baseline to test your ideas.

\subsection{Our Contributions}

We test our ideas by implementing an LTH-based approach to pruning on the SST-2 \cite{socher2013recursive}, STS-B \cite{cer2017semeval}, and CoLA \cite{warstadt2018neural} datasets which are very well-known datasets in the GLUE \cite{wang2018glue} NLP benchmark. We thus built Cam-Cut based on GradCAM, Smooth-Cut based on SmoothGrad, and Smooth-Cam-Cut which combined the ideas of both approaches. For measuring size reduction, we use the Compression Ratio, which is the ratio of the number of parameters in the unpruned model to the number of parameters in the pruned model. For benchmarking our results against the baselines, we used ShrinkBench \cite{ortiz2019standardizing} which provides a standard implementation for pruning the fully-connected layers and convolutional models.

In this work, our contribution is five-fold:

\begin{itemize}
\item Cam-Cut: A GradCAM-based pruning metric that performs well, especially on high compression ratios.
\item Smooth-Cut: A SmoothGrad-based pruning metric that performs well with noisy gradient signals.
\item We outperform the baselines at different compression ratios on three GLUE datasets and discuss the results and effectiveness of our strategies.
\item We compare the pruning masks for Smooth-Cut and Cam-Cut with the pruning masks for our baselines to show that these strategies identify different parameters for pruning.
\item We also analyze our pruning masks to give important insights into BERT layers and parameters thus allowing a deeper understanding of the BERT model.
\end{itemize}

\section{Problem Formulation}

We aim to prune BERT-based models, with task-specific classification and regression layers over the BERT pooling layer for the SST-2, STS-B, and CoLA datasets. As we perform pruning, our inputs include the training dataset, the BERT model with a regressor or classifier layer, and our pruning strategy. The output is a set of pruning masks over every parameter in the model that carries information on which weights and biases are retained and which ones are pruned. We reinitialize the network parameters with pre-pruned values, and fine-tune the models with the mask applied, thus obtaining our final performance of the pruned model. Currently, we are performing single-shot unstructured pruning using the LTH for Cam-Cut, Smooth-Cut, and SmoothCam-Cut strategies. The key challenges we address involve augmenting GradCAM as Cam-Cut, and SmoothGrad as Smooth-Cut using LTH and applying them to multi-batch, multi-token sequences as interpretation methods were used only for single sequences.

We have included the standard baseline results of LayerMagWeight and GlobalMagWeight strategies as proposed in the LTH paper. To the best of our knowledge, we did not find any relevant works in single-shot unstructured pruning that provide a comprehensive review of pruning strategies for large language models. In addition, the current state-of-the-art in iterative unstructured pruning, Optimal BERT \cite{kurtic2022optimal} also report their single-shot performance. However, we can not compare our results with this work as they only provide GPU-specific results for 2-out-of-4 sparsity on NVIDIA Ampere GPU \cite{mishra2021accelerating}. For this, they use magnitude-based pruning strategies for benchmarking their results which is also our baseline.

\subsection{Baseline Models}
%Describe the baseline system in some detail. Use a figure. Describe the components of this baseline in detail.

\subsubsection{Global Magnitude Pruning}
The Global Magnitude Pruning method or GlobalMagWeight method is originally proposed in the LTH paper. It uses the importance score $S$ which is computed using the network parameters $w$ as:

\begin{equation}
   S = |w| 
\end{equation}

The importance score is then converted into a mask by selecting parameters with the maximum score from throughout the architecture, irrespective of the module or layer they are in. The mask is a binary tensor for every parameter tensor with 1 for all included weights and 0 for excluded ones. Thus, the mask creation for a parameter $w$ is done by:

\begin{align}
    mask[w] = 
        \begin{dcases*}
            1, & if $w \in $ top $q$ quantile \\
             &  of the parameters overall,\\
            0, & otherwise. 
        \end{dcases*}
\end{align}
where $q$ is determined using the compression ratio and a number of prunable parameters throughout the model.

\subsubsection{Layer Magnitude Pruning}
The Layer Magnitude Pruning method or LayerMagWeight method was originally proposed in the LTH paper as well. It uses the importance score in the same way as GlobalMagWeight. However, unlike the global magnitude implementation, the importance score is converted into a mask by selecting parameters with the maximum score from that particular module and not the entire network. The mask is a binary tensor for every parameter with 1 for all included weights and 0 for excluded ones. Thus, the mask creation for a parameter $w$ is done by:

\begin{align}
    mask[w] = 
        \begin{dcases*}
            1, & if $w \in $ top $q$ quantile of\\
             &  parameters in given tensor,\\
            0, & otherwise. 
        \end{dcases*}
\end{align}
where $q$ is determined using the compression ratio and the number of prunable parameters in the considered module.

\subsection{Challenges Faced}

\subsubsection{Issues with baseline approaches} 
As seen above, the baseline methods only consider the magnitude of the weights for pruning. For zero-shot pruning, they are completely task-agnostic as the model is trained only for language modeling. However, for single-shot pruning, the weights do reflect some significance of the task at hand due to fine-tuning but the pruning metric does not reflect the same directly. We thus look into gradient-based pruning methods. One of the standard implementations in the ShrinkBench framework called LayerMagGrad used parameter magnitude and gradient as a metric as shown for the given parameter $w$:

\begin{equation}
    S_{LayerMagGrad} = |w \times \dfrac{\partial{L}}{\partial{w}}|    
\end{equation}

This is similar to the Vanilla Gradient \cite{simonyan2013deep} formulation in interpretability literature. Here, the gradients are known to be noisy which affects the pruning performance. In addition, these gradients are also not very good estimators of their local neighborhoods. We also verified this in our tests as only gradient-based scoring did not perform well whereas a combination of weights and gradients performed better than the baseline of only weights. A major observation was that till the compression ratio of 2, all methods performed similarly, and major gains due to gradient information are observed as we go from compression ratios of 2 to 4.

% \subsubsection{Issues with baseline implementation}
% We also observed several implementation challenges in the ShrinkBench framework which made it unsuitable for transformer architectures unless updated. The GlobalMagWeight implementation was not suitable for very large compression ratios as it would completely remove many weights from a single layer. This could cause problems in the information flow in the transformer structure. The transformer architecture has many modules like attention which have sub-modules and ShrinkBench did not support pruning in parent modules not having parameters. The LayerMagGrad implementation considered gradients over only a single batch, which provided poor results and thus excluded from our baselines. We have corrected the same and accumulated the importance scores derived from gradients over multiple batches. This clearly helps in more complex tasks like the STS-B dataset as shown later in our results.

\section{Implementation Details}

\subsection{Dataset Processing}
To handle inputs for the three datasets: SST-2, STS-B, and CoLA, we use dataset-specific tokenization techniques. As SST-2 and CoLA are single text classifications, we directly tokenize them without introducing different masking and separator tokens, whereas, in the STS-B dataset, regression is done according to pair of sentences. These are separated by a separator token and different segment embeddings are provided. For STS-B regression we use sigmoid as the last layer and multiply it with $5$ to make it suitable for the $5$-level similarity regression.

\subsection{Gradient Computations}
All our proposed methods need a gradient of the loss with respect to the considered weight for importance and mask calculations. Thus, for computing the gradients, we break the entire training set into multiple batches and one gradient is computed per batch. This gradient is converted into the importance metric according to the selected strategy and the importance scores are accumulated over the entire dataset. This ensures that gradients over the entire dataset are considered and not over a random batch. For reducing computation, we consider the first $1,000$ batches generated from the dataset.

\subsection{Prunable Layers}
Currently, not all parameters of the BERT model are considered prunable. The Embedding layers are generally not considered for pruning in various works thus we have also not pruned them. Similarly, LayerNorm layers also have parameters for normalizing outputs which are not pruned. The classifier and regressor layers making the final predictions are also not being pruned currently. Thus, out of 110 million BERT parameters, only 85 million are considered prunable.

\subsection{Cam-Cut (LayerGradCAMShift)}

\begin{figure}[ht!]
    \centering
    \includegraphics[width=0.5\textwidth]{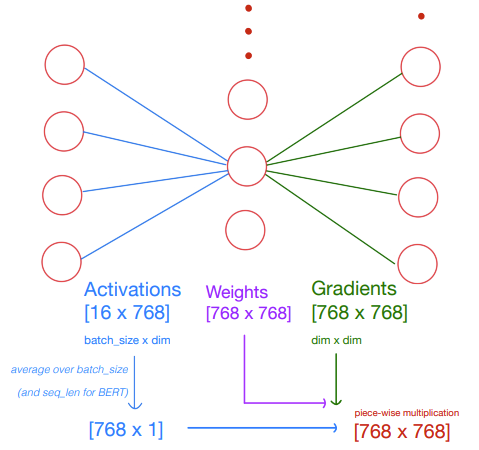}
    \caption{GradCAM for a 3-layer MLP. Here, we average the activations over the current batch to obtain a single vector, which we then multiply piece-wise with each column of the gradient matrix. The resultant matrix is then piece-wise multiplied by the magnitude of the weights, and after taking the absolute value of each element, we obtain its importance for the given batch.}
    \label{fig:gradcam}
\end{figure}

\subsubsection{Overview}
As shown in \ref{fig:gradcam}, the LayerGradCAM approach involves 3 separate ideas used together to get good performance in pruning. The Layer part indicates that the method is layer-wise pruning and thus prunes the same quantile from every tensor. The Grad part indicates that it uses the gradients averaged over the entire dataset as described previously for computing the weight scores. CAM stands for Class Activation Map, an idea borrowed from CNNs where the final representation of all convolutions was considered as a much smaller image indicating the importance of certain regions of the image when rescaled back to the size of the image. This idea directly translates to transformer architectures where every token gets its own representation at every layer, thus every layer has its own class activation map and no rescaling is needed. The Shift is a correction for activation functions like GELU and Relu which will be described further in \ref{shift} and introduced to significantly enhance performance.

\subsubsection{Restructuring GradCAM}
A major challenge arises while translating GradCAM to transformers in the above formulation. As every weight needs a single definite score for pruning, but every token and batch generates multiple outputs for the same layer. Thus, we average pool the outputs before the application of the activation function over all sequences and tokens for a given batch.

\subsubsection{Introducing Shift}
\label{shift}

\begin{figure}[ht!]
    \centering
    \includegraphics[width=0.5\textwidth]{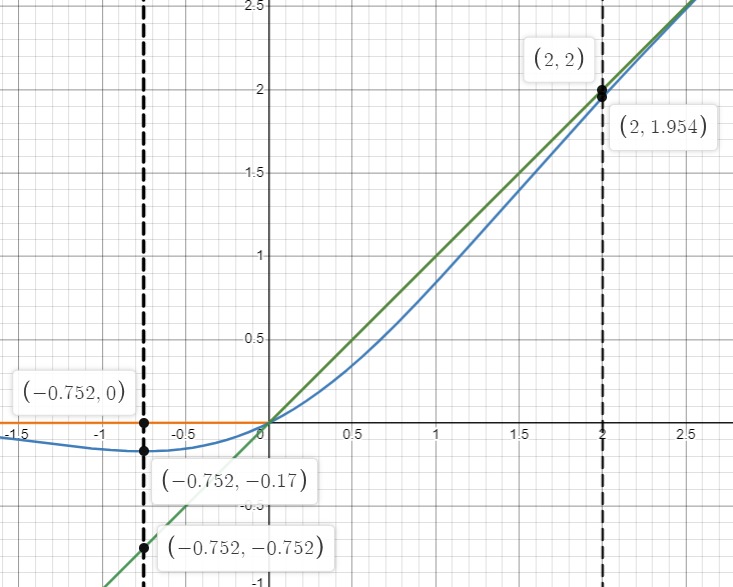}
    \caption{The black lines indicate the unactivated outputs of 2 neurons being considered for CAM. If Shift is introduced, the linear function gives the maximum difference between activated and unactivated neurons when compared to GeLU and ReLU. It also makes the sign of deactivated output irrelevant as the absolute function focuses more on the magnitude of the function.}
    \label{fig:reluGELU}
\end{figure}

The BERT architecture uses the GELU activation function for activating neurons but it is treated as a different module while pruning. Thus, the CAM outputs obtained are not activated. This results in the LayerGradCAM model performing significantly worse as seen in our SST-2 results \ref{fig:sst2}. We then introduced the ReLU activation function which is very similar to the structure of the GELU activation function used in BERT. However, this causes many scores to become zero and results in the generation of erroneous masks. Thus, to ensure that CAM scores directly indicate the likelihood of a neuron being activated, we shifted all scores up by a scalar $\lambda$ which is a hyper-parameter. We believe this helped very clearly discern importance even among deactivated neurons thus giving a very good performance across our 3 datasets. For our experiments, we used a value of $10$ for $\lambda$.

\begin{figure*}
    \centering
    \includegraphics[width=0.6\textwidth]{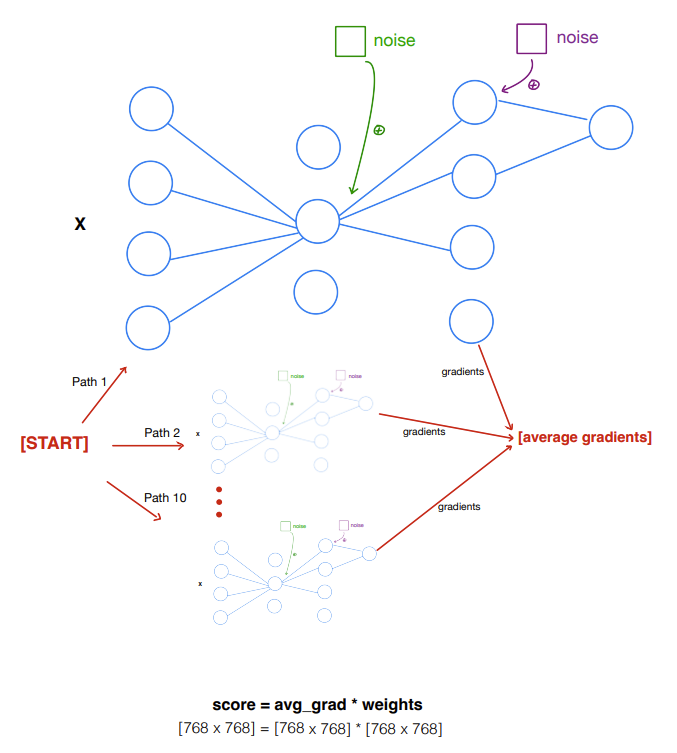}
    \caption{A depiction of smooth grad: Starting with an input vector, we add random noise samples from Gaussian distribution at each step of the forward process, and we do this for a total of 10 paths. The final gradient of the network for this input will be the average gradient across all the paths.}
    \label{fig:smoothcam}
\end{figure*}

\subsubsection{Final Scoring metric and Mask}
Taking cumulative of all our ideas we thus derive the final scoring metric for LayerGradCAMShift as:

\begin{equation}
    S = | w \times \frac{\partial{L}}{\partial{w}} 
        \times (\frac{1}{l_b}\sum_{l_b} \frac{1}{l_s}\sum_{l_s}(w \cdot x + b) + \lambda)|
\end{equation}

where $l_s$ is the sequence length, $l_b$ is the batch length, and $\lambda$ indicates the shift.

The final mask can be obtained like LayerMagWeight:

\begin{align}
    mask[w] = 
        \begin{dcases*}
            1, & if $w \in $ top $q$ quantile of\\
             &  parameters in given tensor,\\
            0, & otherwise. 
        \end{dcases*}
\end{align}
where $q$ is determined using the compression ratio and the number of prunable layers.

\subsection{Smooth-Cut (LayerSmoothGrad)}

\subsubsection{Need for Smoothing}
In interpretability literature, it was commonly observed that Vanilla Gradients or using only gradients provided noisy results thus normal noise was added to the input and the weighted sum of the gradients of noisy input was considered as a mask. The probability of the noise was considered as the weight thus making larger noise weigh less and smaller noise weigh more. This created a smoothing effect over the piecewise-linear manifold which provides a much better approximation of the gradients. This effect is clearly depicted in \ref{fig:linearpiece} Thus, we can formulate this in terms of weights, input, and biases as follows:

\begin{figure}[ht!]
    \centering
    \includegraphics[width=0.4\textwidth]{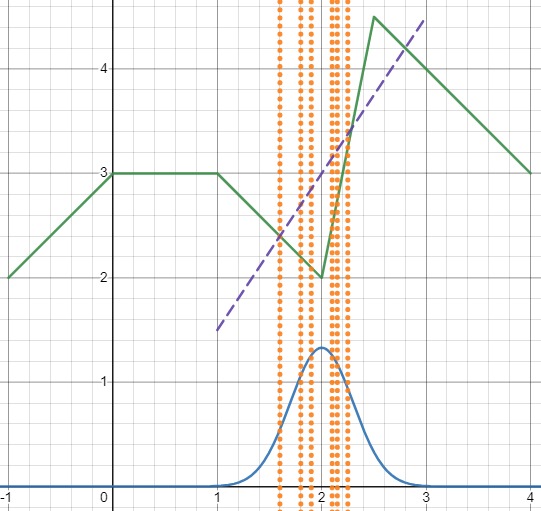}
    \caption{Smoothing Gradients of a piecewise linear function using sampling of noisy input}
    \label{fig:linearpiece}
\end{figure}

Let $z$ be a random variate sampled from a normal distribution with a variance of 0.01 and a mean of 0.

\begin{equation}
    \begin{aligned}
        w_{noisy} &= w + z   \\
        b_{noisy} &= b + z   \\ 
        y_{noisy} &= (w + z) \cdot x + (b+z) \\
                  &= (w\cdot x + b) + z \cdot ( 1 + x) \\
    \end{aligned}
\end{equation} 

As the output of every layer is much smaller than $1$ in our observation we approximated this as:

\begin{equation}
    y_{noisy} \approx (w\cdot x + b) + z    
\end{equation}

This is done using a forward hook in every layer and the $backward()$ function in $PyTorch$ is used to sample the gradient. For ease of implementation and as the variance in noise is small, we have taken a numeric average instead of the weighted average, unlike the SmoothGrad paper.

\subsubsection{Implementation at scale}
Ideally, every parameter tensor must be sampled with noise while keeping all other parameters constant to get a good estimate but this is not scalable with architecture as large as BERT. To overcome this limitation we instead reduce the variance in noise as it will be accumulated over multiple layers and add it in the forward pass at every layer. As the linear combination of Gaussian normal noise is also Gaussian, and the variance in noise is small we believe our method is able to sample the locality of weights very well. This thus gives a correlated noisy estimate of gradients for all parameters in every forward propagation. We thus sample $\eta$ such paths and average the gradients over all these paths giving a smoothed gradient approximation. As we sample $\eta$ different paths for every input, this method needs $\eta$ iterations for every sample whereas LayerGradCAMShift needs only $1$. To scale this approach to larger datasets, and fairly evaluate the two methods, we run all Smoothing methods with only 100 batches of gradients and each batch undergoes 10 smoothing operations denoted by $\eta$. This is clearly shown in \ref{fig:smoothcam}.

\subsubsection{Final Scoring metric and Mask}
Taking cumulative of all our ideas we thus derive the final scoring metric for LayerSmoothGrad as:

\begin{equation}
    S = |w \times \frac{1}{\eta} \sum_{i=1}^{i=\eta} \frac{\partial{L_{Noisy}}}{\partial{w}} |
\end{equation}

The final mask can be obtained like LayerMagWeight:
\begin{align}
    mask[w] = 
        \begin{dcases*}
            1, & if $w \in $ top $q$ quantile of\\
             &  parameters in given tensor,\\
            0, & otherwise. 
        \end{dcases*}
\end{align}
where $q$ is determined using the compression ratio and the number of prunable parameters. \\

% State the high-level principles behind your ideas. Then describe each idea in detail. In this section, you should provide enough detail
% of the system so that someone else if they want to implement your system has enough information to replicate it. Use a figure if appropriate.

%[Note: Name the ideas with descriptive terms. I am just listing them as 1, 2, and 3.]

\section{Evaluation}

\begin{figure*}[htbp!]
    \centering
    \includegraphics[width=0.8\textwidth]{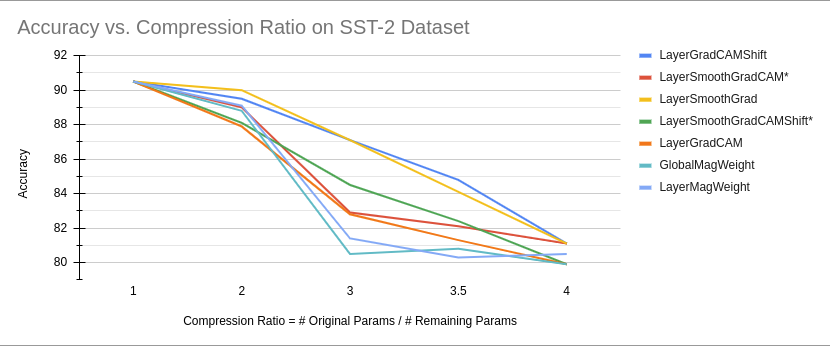}
    \caption{SST-2 dataset}
    \label{fig:sst2}
    \end{figure*}
    \begin{figure*}
    \centering
    \includegraphics[width=0.80\textwidth]{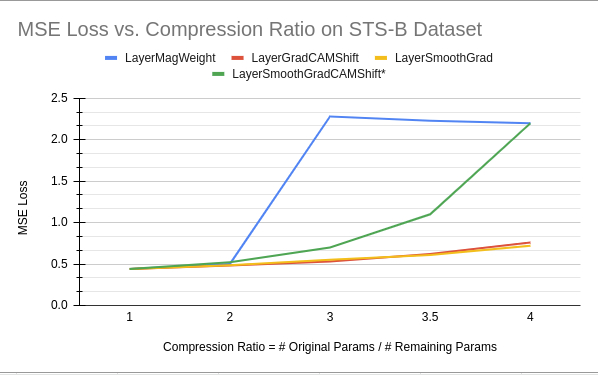}
    \caption{STS-B dataset}
    \label{fig:stsb}
    \end{figure*}
    \begin{figure*}
    \centering
    \includegraphics[width=0.8\textwidth]{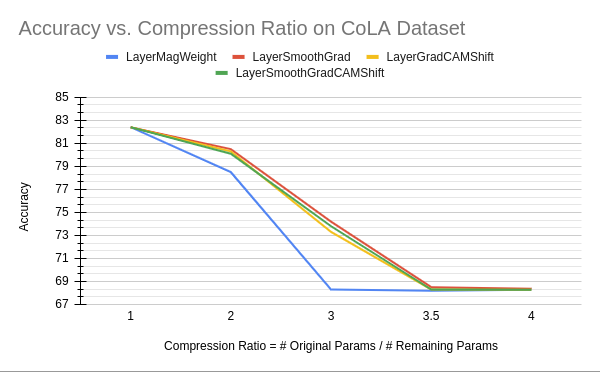}
    \caption{CoLA dataset}
    \label{fig:cola}
\end{figure*}
%State the purpose of your evaluation. How should one evaluate a system for your task. What are the questions to ask?

The main purpose of this evaluation is to check how much performance is retained for the given dataset and compression ratio, thus giving us deeper insights into the applicability and differences in our pruning methods.

To evaluate our pruning methods we test them over 3 datasets, SST-2, STS-B, and CoLA, and for $4$ compression ratios of $2,3,3.5,4$. We observed that there wasn't much variation between ratios of $1$ and $2$, thus they were not run for all techniques.

The major questions we ask during evaluation are:
\begin{enumerate}
\item How much performance is lost for different compression ratios across different pruning strategies?
\item How important are parameters at various pruning thresholds thus giving an idea if performance behaves like the Pareto principle? 20\% weights can give over $80\%$ of the expected results.
\item How different are the pruning masks generated by the various strategies and how do these differences turn out across different layers?
\end{enumerate}

\subsection{Dataset Details}
The following GLUE datasets were used:

\begin{enumerate}
    \item SST-2 (Stanford Sentiment Analysis Task): This has $67k$ training samples with 1.8k test samples. It allows only $2$ sentiment classes for classification.
    \item STS-B (Semantic Textual Similarity Benchmark): This has $7k$ training samples with $1.4k$ test samples. It has a continuous target score ranging from $0$ to $5$ where $0$ indicates opposite sentences and $5$ equivalent sentences. This is a regression problem.
    \item CoLA (Corpus of Linguistic Acceptability): This has $8.5k$ training samples and $1k$ testing samples. It tests if a sentence is grammatically acceptable or not and has a single sentence as input.
\end{enumerate}

\subsection{Evaluation Measures}
We analyze the performance in terms of accuracy for SST-2 and CoLA and Mean Squared Error for the STS-B dataset and how it varies with increasing compression ratios in our datasets.

\subsubsection{Evaluation Process}
The pruning process involves a set of fine-tuning epochs which are followed by pruning and then re-fine-tuning. A batch size of $16$ with a learning rate of $1e-5$ and the Adam optimizer was used for SST-2 and CoLA datasets. For the STS-B dataset, we used a learning rate of $2e-5$. For the SST-$2$, dataset $3$ initial fine-tuning epochs were run and $8$ final fine-tuning epochs were used. For CoLA $5$ initial fine-tuning epochs and $12$ final-fine-tuning epochs were used. Finally, for the STS-B dataset, we used $8$ initial fine-tuning epochs and $10$ final-fine-tuning epochs. The models were finally chosen using early stopping and the learning rate, batch size, and the number of epochs were selected after observing the loss over multiple runs.

% \subsection{Baselines}
% The LayerMagWeight and GlobalMagWeight models are considered our baselines. The implementation of these models is done according to the scoring functions and mask derivations described previously.

\subsection{Results}
We first present our results on the SST-2 Dataset in \ref{fig:sst2}. Here we observe that we significantly outperform the baselines for all compression ratios. As can be observed, up to the compression ratio of $3$ LayerMagWeight outperforms GlobalMagWeight but both have similar performance after that. We also observe that Shift is essential for LayerGradCAM to perform well and the performance of LayerGradCAM is barely at the baseline possibly due to the weight signal. Another important observation is that even though LayerSmoothGrad and LayerGradCAMShift work well, their combinations LayerSmoothGradCAM and LayerSmoothGradCAMShift do not appear to perform equally well. We later identified that these combination methods were using gradients of only $1$ batch and thus do not seem to perform as well. We corrected this in CoLA and have provided the results.

\begin{figure*}[ht!]
    \centering
    \subfloat[Mean,Compression=3]{\includegraphics[width=0.35\textwidth]{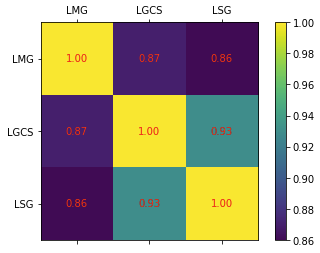}
    \label{mean3}}
    \hfil
    \subfloat[Min,Compression=3]{\includegraphics[width=0.35\textwidth]{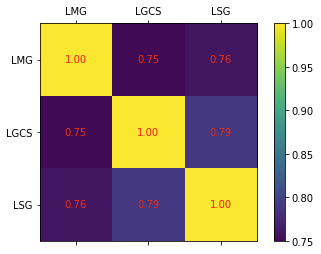}
    \label{min3}}
    % \caption{At compression=3.0} 
    \label{fig2}
    \centering
    \subfloat[Mean,Compression=3.5]{\includegraphics[width=0.35\textwidth]{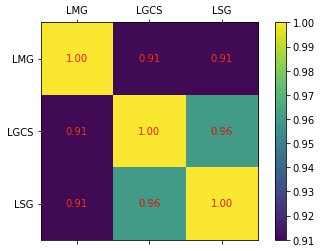}
    \label{mean35}}
    \hfil
    \subfloat[Min,Compression=3.5]{\includegraphics[width=0.35\textwidth]{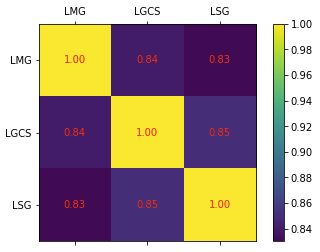}
    \label{min35}}
    \caption{Here, \ref{mean3} and \ref{min3} represent the confusion matrix of the mean and min values across all layers, respectively, for a compression ratio of 3. \ref{mean35} and \ref{min35} represents the confusion matrix of the mean and min values across all layers, respectively, for a compression ratio of 3.5. \textbf{LMG}: LayerMagGrad, \textbf{LGCS}: LayerGradCAMShift, \textbf{LSG}: LayerSmoothGrad.} 
    \label{fig9}
\end{figure*}
For the STS-B dataset in \ref{fig:stsb}, the LayerGradCAMShift and LayerSmoothGrad methods perform extremely well incurring very low losses even at a compression ratio of $4$ whereas the LayerMagWeight baseline drastically loses its performance from a compression ratio of $3$ onwards.

For the CoLA dataset \ref{fig:cola} we observe similar trends but possibly due to  dataset specific reasons the performance flattens at $3.5$ compression at about $68.3$. In SST-2 and STS-B there was a much larger variation at compression ratios of $3.5$ and $4$ compared to CoLA.

Even at compression ratio $2$ across all datasets the performance is very similar irrespective of the pruning method used. At ratios of $3$ and $3.5$, the difference increases significantly and at the compression ratio of $4$, where less than $5\%$ of prunable parameters remain, we see the performance of all methods tending to converge at least for SST-2 and CoLA datasets in \ref{fig:sst2} and \ref{fig:cola}.

%Main results that compare your ideas to the baselines.
%What does this result tell you?

\subsection{Analysis}
As observed in our results the performance drops steeply with the compression ratio unless the method of pruning is suitable. Thus we conclude that it is essential to choose the right pruning strategy to get good results, especially for compression ratios greater than $2$. For all $3$ datasets, the performance at compression ratio $2$ varies marginally across all pruning strategies. We believe these results are due to the reason that many sub-networks can perform very well unless they are constrained to be too small. This is in accordance with \cite{prasanna2020bert}.

\begin{figure*}[htbp!]
    \centering
    \subfloat[Compression=3]{\includegraphics[width=0.35\textwidth]{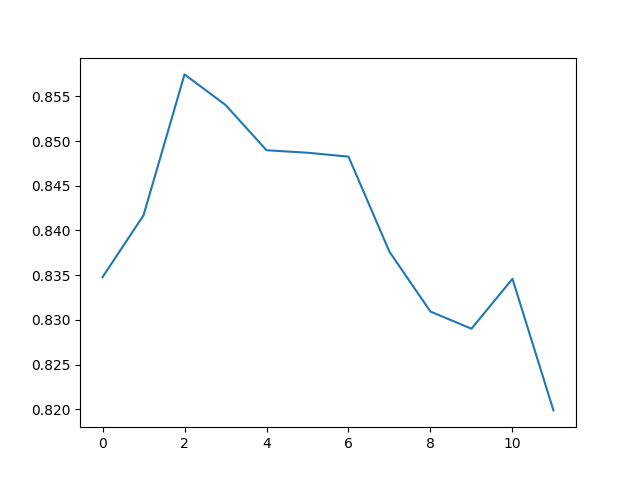}
    \label{fig121}}
    \hfil
    \subfloat[Compression=3.5]{\includegraphics[width=0.35\textwidth]{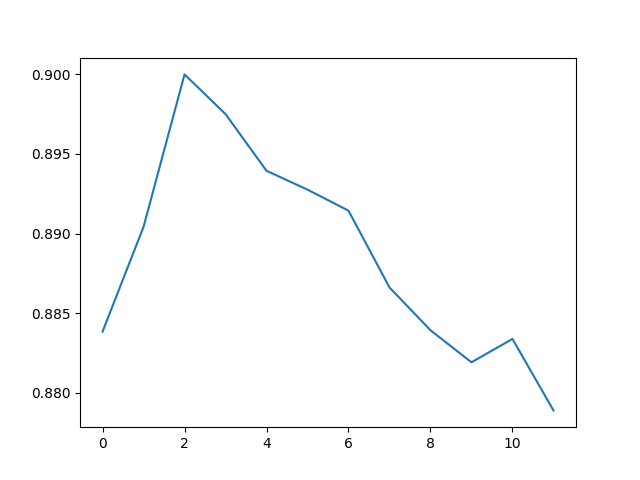}
    \label{fig101}}
    \caption{Cam-Cut v/s LayerMagWeight attention-head IOUs at two different compression levels} 
    \label{fig10}
\end{figure*}
\begin{figure*}[htpb!]
    \centering
    \subfloat[Compression=3]{\includegraphics[width=0.35\textwidth]{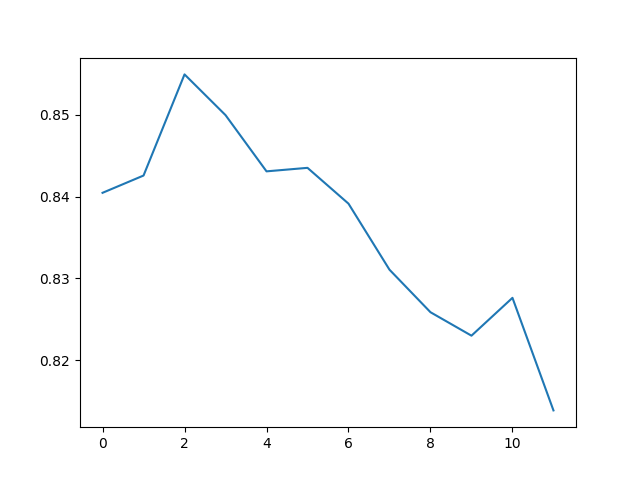}
    \label{fig211}}
    \hfil
    \subfloat[Compression=3.5]{\includegraphics[width=0.35\textwidth]{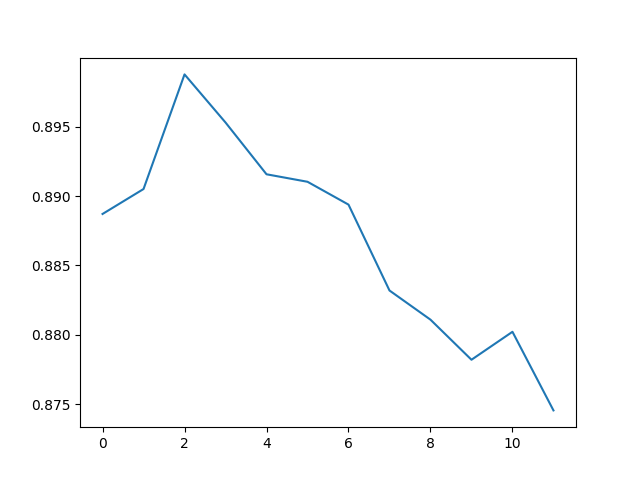}
    \label{fig111}}
    \caption{Smooth-Cut v/s LayerMagWeight attention-head IOUs at two different compression levels} 
    \label{fig11}
\end{figure*}
\begin{figure*}[htpb!]
    \centering
    \subfloat[Compression=3]{\includegraphics[width=0.35\textwidth]{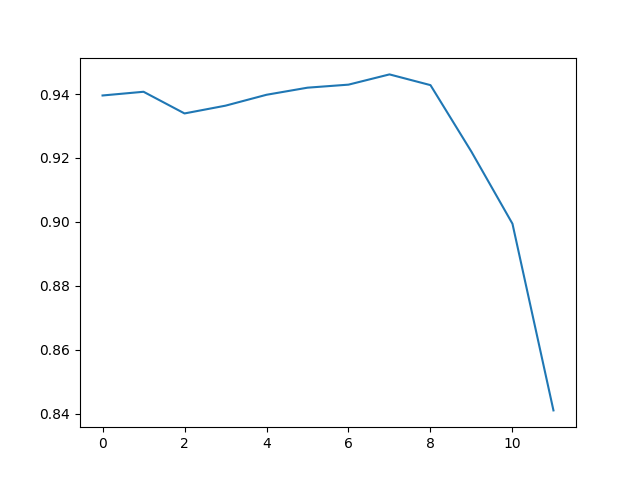}
    \label{fig21}}
    \hfil
    \subfloat[Compression=3.5]{\includegraphics[width=0.35\textwidth]{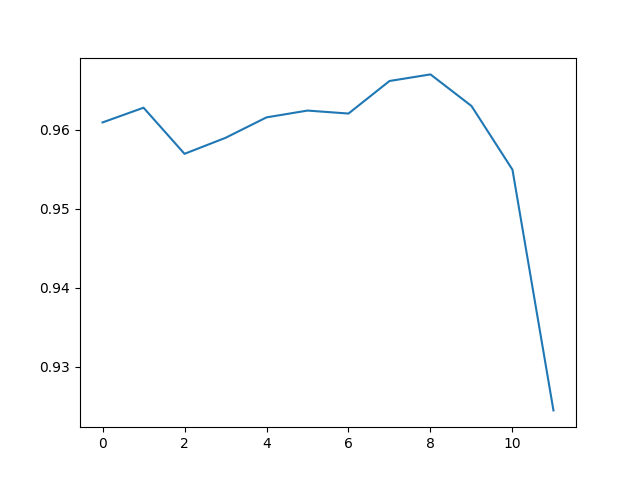}
    \label{fig23}}
    \caption{Cam-Cut v/s Smooth-Cut attention-head IOUs at two different compression levels} 
    \label{fig12}
\end{figure*}

Simultaneously we also observed that the weight information present in all our metrics is also very important. We did some experiments initially with compression ratios of $2$ without including the weight term and observed that the performance of the LayerGradCAM was not even surpassing the baseline. Thus we have included the weight term in all the strategies.

Beyond the compression ratio of $2$, we find that gradient information becomes extremely important. This is very evident as gradient-based methods easily surpass only weight-based ones. This is not seen until the compression ratio of $2$ and not even in some of our experiments with compression ratios of $1.5$.

Even in gradient-based methods, it is very important to have a gradient over a large significant portion of the dataset. Our runs using the combined methods LayerSmoothGradCAMShift and LayerSmoothGradCAM were initially taking gradients over only a single batch of 16 samples thus giving the degraded performances as seen in SST-2 and STS-B. But with many more batches, the performance improves significantly as seen in the CoLA dataset.

We also perform a detailed analysis of the masks created by the methods LayerMagWeight, LayerGradCAMShift, and LayerSmoothGrad for SST-2 dataset. We find the Jaccard distance of these masks which represents the intersection over the union of the weights which are selected after pruning at compression ratios of $3$ and $3.5$. We have computed the mean and min IOU across all layers for every pair of strategy as shown in \ref{fig9}. We can observe that the mean and min value of LayerMagWeight with both, Cam-Cut and Smooth-Cut is low, and the same values when we compare Cam-Cut and Smooth-Cut are high. This gives a hint at the similar and superior performance of both these strategies.

We further analyze these mask differences across layers of BERT as depicted in \ref{fig10}, \ref{fig11}, and \ref{fig12}. We find that LayerMagWeight has IOUs of about 82\% to 86\% across layers and this tends to drop as we go higher in the architecture irrespective of the strategy. As this trend is most pertinent with LayerMagWeight, it implies that the shallower and middle layers are the ones storing a lot of language modeling information in their weights and their weights are more important compared to gradients. On the other side, the higher weights seem more task-based, especially in the final layer where even LayerGradCAMShift and LayerSmoothGrad have more differences than others. Simultaneously, we also observe that LayerSmoothGrad and LayerGradCAMShift have very similar masks throughout various layers. On average they have about 95\% similarity until the final layer of the model. This is a strong indicator that it is extremely likely that both models are converging to very similar sub-models despite having very different scoring metrics and structures. This is also probably the reason for both methods giving very similar performance. This also shares parallels with the interpretability literature where saliency masks are better if they are sparse and highlight the features of the region of interest. Every pixel highlighted must explain more variance in the output class than every pixel not highlighted. Similarly, every parameter selected must provide a greater gain in performance than every parameter removed.

% \begin{enumerate}
% \item This is where you tell me what worked and what didn't work.
% \item An error analysis is key. 
% \item If you make any guesses as to why something doesn't work try to find evidence to support it. (E.g.. If you say our CNN model isn't able to handle long sentences well, then plot the performance against sentence length and see if this is true).
% \end{enumerate}

% \subsection{Code}

% The code for this work with a detailed README.md is available at: \url{https://github.com/RuskinManku/DeepCuts}
% % Please provide a {\bf google drive link to your packaged code} or give us a link to your github repo hosting these.  

% The code should be structured with a README that clearly specifies the following:

% \begin{enumerate}
% \item List the original source for your code base. Include the URL to the original source. 
% \item The list of files that you modified and the specific functions within each file you modified for your project. 
% \item A list of commands that provide how you train and test your baseline and the systems you built. 
% \item A list of the major software requirements that are needed to run your system. (E.g. Tensorflow 2.3, Python 243.12, CUDA abd2.0, nltk-2401.11, allen-nlp 5.0). 
% \end{enumerate}

% These descriptions should be adequate enough to help anyone who wants to run your system.

\section{Conclusion and Future Work}
In this paper, we extended interpretability techniques to pruning and observed that many ideas developed to enhance interpretability transfer well to pruning. This also indicates that other interpretability works like DeepLIFT \cite{shrikumar2017learning} can also be extended to further improve results. Our analysis also indicates that very different pruning metrics may end up converging to the same lottery ticket thus giving more insights into how different parameters play different roles in different downstream tasks. We also plan to analyze the similarity of pruning masks across different datasets and analyze differences in masks across layers of the BERT model. Finally, our insights in pruning metrics can translate into actual parameter reduction through structured iterative pruning in the future.

% We were faced with multiple challenges, from understanding the  existing code base of ShrinkBench, implementing gradient/activation-based methods using hooks in PyTorch, formulating robust models for different tasks at hand, and brainstorming how our ideas can be improved above the baseline, for which we had to dive deep into the BERT architecture and the theory of loss function optimization. Since we had to run many experiments involving combinations of compression ratios and pruning strategies, hyper-parameter tuning was particularly important. We observed how robust pruning techniques work particularly well at high compression ratios, and reading the literature, we realized how pruning during fine-tuning, or pruning at once, in the beginning, using LTH, can produce significantly different results. In all, it's safe to say that the field of pruning requires one to think about Machine Learning models at the scale of individual weights and neurons, which makes us excited to think how small innovations in this field can lead to robust pruned models, making them accessible for more and more people.

\section{References}

\bibliographystyle{apalike}
\bibliography{deepcuts}

\end{document}